\documentclass[letterpaper, 10 pt, conference]{ieeeconf} 
\IEEEoverridecommandlockouts                              
\usepackage{graphics} 
\usepackage{graphicx}
\usepackage{epsfig} 
\usepackage{amsmath} 
\usepackage{amssymb}  
\usepackage{booktabs}
\usepackage{array}
\usepackage{float}
\usepackage[space]{cite}    
\usepackage[capitalize]{cleveref}  
\usepackage{url}

\title{\LARGE \bf
TacCap: A Wearable FBG-Based Tactile Sensor for Seamless Human-to-Robot Skill Transfer
}

\usepackage[dvipsnames]{xcolor}

\author{Chengyi Xing$^{*1}$, \space  Hao Li$^{*1}$, \space  Yi-Lin Wei$^{2}$, \space Tian-Ao Ren$^{1}$, \space Tianyu Tu$^{1}$, \space Yuhao Lin$^{2}$ \\ \space Elizabeth Schumann$^{1}$ \space Wei-Shi Zheng$^{2}$ \space Mark R. Cutkosky$^{1}$
\thanks{*Equal contribution}
\thanks{$^{1}$The authors are with Stanford University, USA {\tt \{chengyix, li2053, tianao, ttyaaron, ejs, cutkosky\} @stanford.edu}}
\thanks{$^{2}$The authors are with Sun Yat-sen University, China {\tt \{weiylin5, linyh96\} @mail2.sysu.edu.cn, wszheng@ieee.org}}}%

\begin{document}
\maketitle
    \thispagestyle{empty}
\pagestyle{empty}
\begin{abstract}
Tactile sensing is essential for dexterous manipulation, yet large-scale human demonstration datasets lack tactile feedback, limiting their effectiveness in skill transfer to robots. To address this, we introduce TacCap, a wearable Fiber Bragg Grating (FBG)-based tactile sensor designed for seamless human-to-robot transfer. TacCap is lightweight, durable, and immune to electromagnetic interference, making it ideal for real-world data collection. We detail its design and fabrication, evaluate its sensitivity, repeatability, and cross-sensor consistency, and assess its effectiveness through grasp stability prediction and ablation studies. Our results demonstrate that TacCap enables transferable tactile data collection, bridging the gap between human demonstrations and robotic execution. To support further research and development, we open-source our hardware design and software.
\end{abstract}
\section{introduction}
Touch is fundamental to the dexterity of the human hand, enabling complex manipulation tasks with precision and adaptability. Tactile sensing provides essential feedback about object properties such as texture, compliance, and slip, allowing for precise force modulation and adaptive control. Without this rich sensory input, even simple manipulations become significantly more challenging.

In robotics, achieving human-level dexterity has been a long-standing challenge. Researchers have explored data-driven approaches, combining multiple sensory modalities to enable robots to tackle various tasks \cite{lee2020making, li2023see, gao2023sonicverse, gao2023objectfolder, liu2024maniwav}. Imitation learning, particularly through supervised learning from human demonstrations, has emerged as a key method \cite{duan2017one, brohan2023rt, chi2023diffusion}. Traditionally, teleoperation has been used to teach robots, but it is often slow, costly, and limited by the mechanical constraints of physical robot setups.

Recent advancements, such as the Universal Manipulation Interface (UMI) \cite{chi2024universal} and DexCap\cite{wang2024dexcap}, offer new solutions by using wearable and portable devices for in-the-wild data collection from human demonstrations. These systems allow human operators to interact directly with the environment, making data collection faster, more scalable, and versatile across a variety of tasks and settings. However, despite their advantages, they lack the ability to capture tactile feedback, which is crucial for replicating the nuanced control required for truly dexterous manipulation.

To address this gap, a promising approach is to equip human operators with tactile sensors, such as tactile gloves, during data collection \cite{sundaram2019learning, delpreto2022actionsense, luo2024adaptive}. However, they present several challenges: they are often complex and difficult to fabricate, prone to durability issues, and susceptible to interference from electromagnetic fields and metallic objects, requiring additional shielding. Moreover, these gloves are designed to conform to the human hand, which creates a gap for skill transfer. 

\begin{figure}[t]
      \centering
      \includegraphics[width=0.95\linewidth]{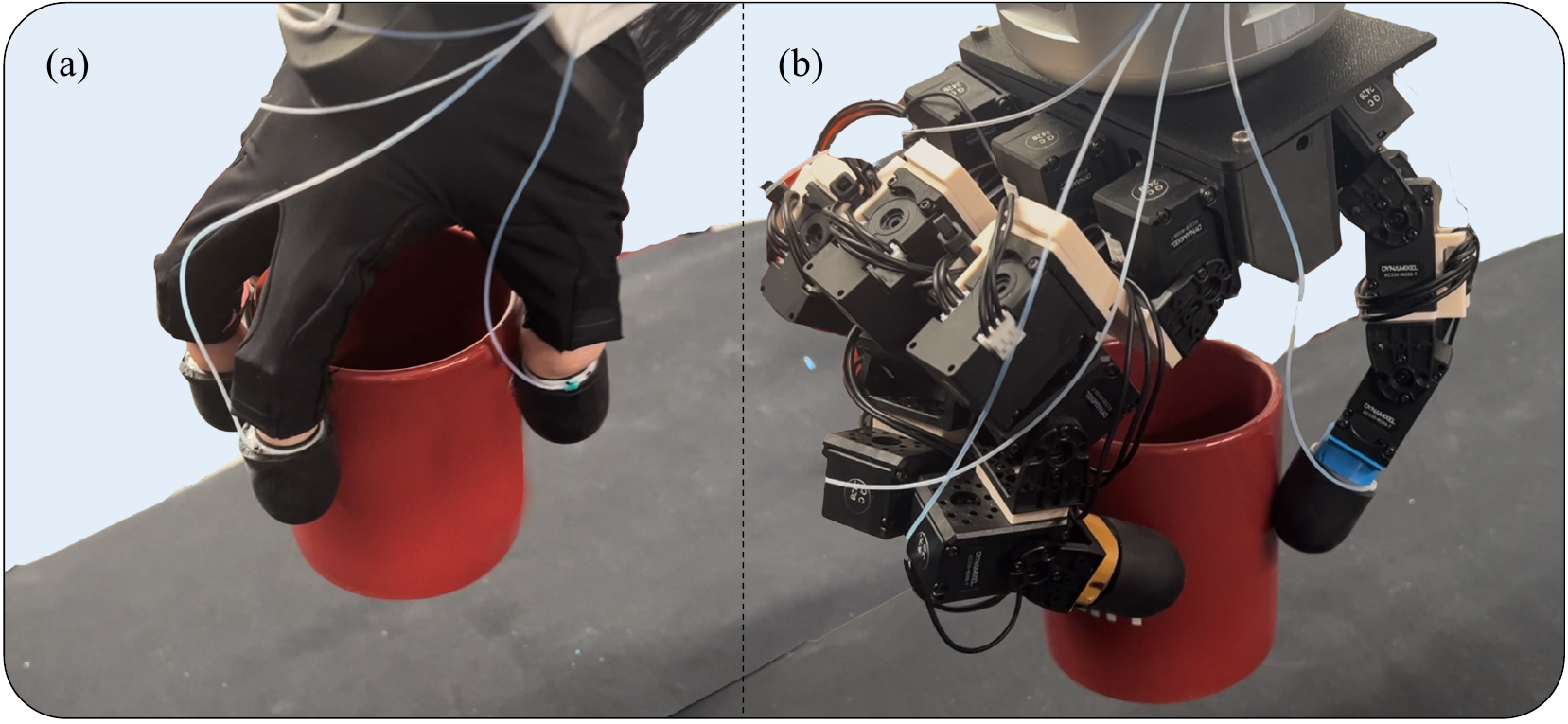}
      \caption{(a) A human operator wearing TacCap tactile sensors to collect demonstration data. (b) A robot hand equipped with the same sensors executing tasks with tactile feedback.}
      \label{demo}
\vspace{-2em}
\end{figure}

\textbf{Contributions}: In response to these challenges, we propose TacCap, a novel wearable tactile sensor with minimal transfer gap between human demonstrations and robot execution. TacCap features a soft thimble structure that can be applied to both human and robot fingertips, enabling the capture of tactile sensing data during human manipulation.

TacCap is based on Fiber Bragg Grating (FBG) technology, which is inherently resistant to environmental factors such as light, water, and magnetic fields. FBGs function as highly sensitive optical strain gauges, capable of resolving strains as small as $10^{-5}$. When strain is applied to the FBGs, it induces shifts in the reflected optical wavelengths, which are detected by an optical interrogator. The interrogator, which can be portable and powered by a battery pack, enables real-time monitoring and can be conveniently carried by a human operator. Additionally, FBG technology allows multiple sensors to be integrated along a single optical fiber, each tuned to a slightly different nominal wavelength. This configuration simplifies wiring while enabling the system to sample dozens of FBGs at rates of up to 2 kHz.

TacCap’s lightweight and scalable design consists of a single optical fiber embedded in a 3D-printed structure, making it easy to fabricate and integrate into human and robotic setups. We open-source the design and software at: {\small \url{sites.google.com/stanford.edu/taccap} }

We validate TacCap through contact position prediction and grasp stability prediction experiments, demonstrating that the collected tactile data exhibits low discrepancy between human demonstrations and robot execution. This promotes skill transfer from humans to robots, addressing a key limitation of existing wearable tactile sensing solutions.
\vspace{-1em}
\section{related work}
\subsection{Robot Data Collection}
Learning from human demonstrations has been a fundamental strategy in advancing robotic manipulation. Traditionally, teleoperation systems have played a crucial role in gathering high-quality demonstrations, enabling robots to imitate human dexterity \cite{chi2023diffusion, zhang2018deep, wang2023mimicplay, shaw2023videodex}. Recent advancements, such as ALOHA \cite{zhao2023learning,fu2024mobile} and GELLO \cite{wu2024gello}, have introduced more efficient data collection frameworks. However, conventional teleoperation setups often suffer from high hardware costs, limited scalability, and constraints on task diversity.

To overcome these limitations, researchers have explored alternative methods that leverage human motion capture for data collection. Systems such as UMI\cite{chi2024universal}, DexCap \cite{wang2024dexcap}, ARCap\cite{chen2024arcap}, DART\cite{park2024dexhub} introduce portable solutions that enable large-scale, in-the-wild human demonstrations. These approaches significantly enhance scalability by capturing diverse manipulation behaviors outside controlled lab environments. However, they primarily rely on visual and kinematic data, omitting the tactile feedback that humans naturally use for dexterous interactions.

\subsection{Wearable Tactile Sensors}
Wearable tactile sensors have been developed to capture rich contact information during human demonstrations. Sensorized gloves \cite{sundaram2019learning, delpreto2022actionsense, luo2024adaptive} are a prominent example, providing detailed force and contact data. However, many existing designs rely on resistive or capacitive sensing, which can be affected by fabrication complexity, electromagnetic interference, and mechanical degradation over time. Due to their glove-based design, these sensors capture tactile information directly from the human hand, including the fingers and palm, following human hand kinematics. However, due to structural and kinematic differences between human and robotic hands, directly using human hand data creates a significant transfer gap, making it challenging to apply to robot learning.

Alternatively, fingertip-wearable sensors, such as ThimbleSense \cite{battaglia2014thimblesense, battaglia2015thimblesense, altobelli2016wearable}, offer a more compact and form-fitting approach by integrating force/torque (F/T) sensing for grasp analysis. While ThimbleSense shares a similar form factor with our proposed TacCap sensor, it is primarily designed for studying human grasp forces rather than facilitating human-to-robot skill transfer. Additionally, its design is more complex, involving extensive electronics and wiring, which limits portability and scalability in robotic applications.

In contrast, our work introduces TacCap, an FBG-based tactile sensor designed to be lightweight, robust, and seamlessly transferable between human and robotic hands. By incorporating FBG technology, TacCap enables high-fidelity tactile data collection, facilitating improved human-to-robot skill transfer. 

\subsection{FBG-based tactile sensing}
FBG sensors have been extensively used across various industries due to their precision and reliability. Their applications span structural monitoring in the oil and gas sector \cite{qiao2017fiber}, real-time load measurements in wind energy systems \cite{park2011real}, and biomedical sensing for healthcare technologies \cite{presti2020fiber}. In recent years, robotics has also benefited from FBG technology, particularly in tactile sensing.

Several approaches have explored the use of FBG sensors for force detection. One method involves force sensor arrays that integrate FBGs and transducers to capture distributed normal forces \cite{heo2006tactile}. Another technique focuses on multi-axis tactile sensing, where FBGs are used to measure both normal and shear strains, particularly for applications in space environments \cite{frishman2021multi}. A different study introduces a soft, large-area sensitive skin that employs neural networks to decode FBG signals, enabling precise contact position detection \cite{massari2022functional}. Meanwhile, an alternative design takes inspiration from biological whiskers, adapting FBG sensors for underwater contact detection \cite{li2024whisker}.

\section{Design and Fabrication}
In designing TacCap, our primary goals were to create a tactile sensor that is reproducible, robust, and capable of capturing high-fidelity tactile information without introducing a transfer gap between human demonstrations and robot executions. To achieve this, we focused on utilizing FBG technology due to its intrinsic advantages in precision, environmental resilience, and ease of integration. In the following subsections, we detail our design considerations, the fabrication process, and the systematic optimization of TacCap to ensure it meets these critical performance criteria.
\subsection{Sensor Design}

\begin{figure}[t]
      \centering
      \includegraphics[width=0.95\linewidth]{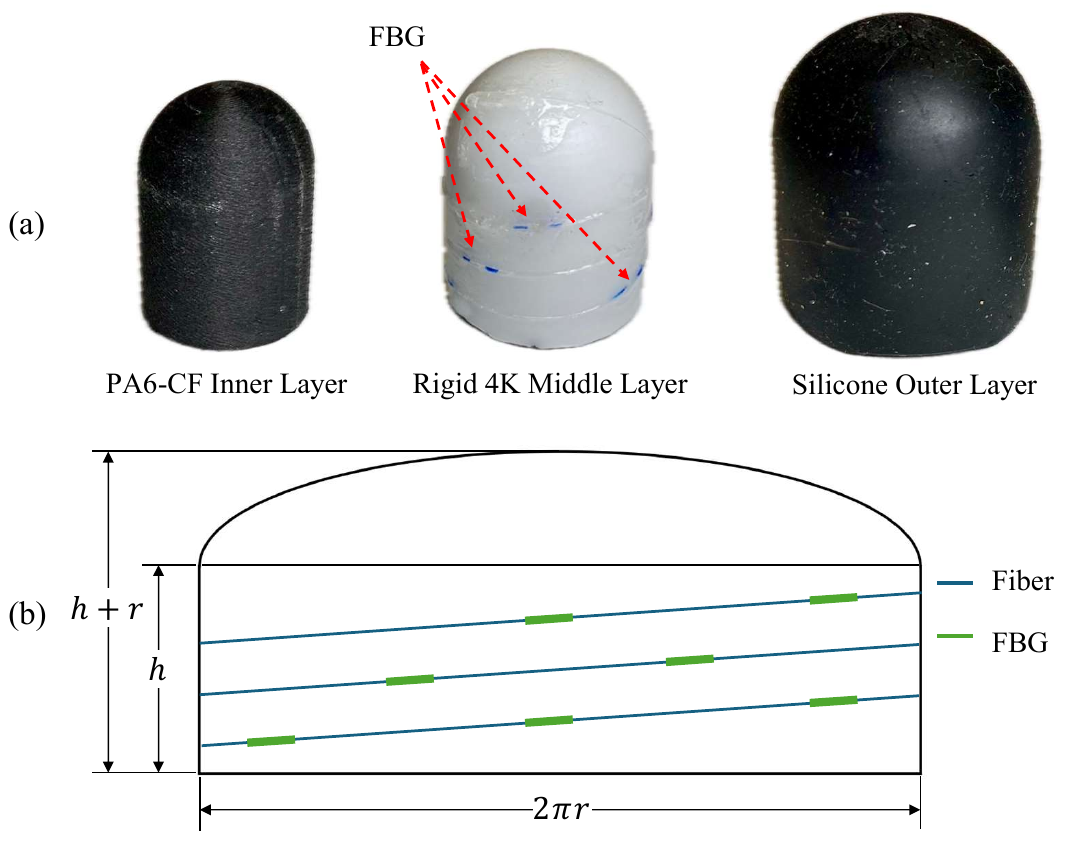}
      \caption{(a) Schematic of the TacCap sensor design and its components. (b) Unwrapped projection of the sensor surface, showing FBG sensor locations.}
      \label{design}
\vspace{-2em}
\end{figure}

Fig.~\ref{design} presents an overview of our tactile sensor design, its components, and the distribution of FBG locations. The sensor consists of three distinct layers: a 3D-printed inner layer (PA6-CF, printed using a Bambu Lab X1C \cite{BambuLab_PA6CF}, accuracy $\approx 0.08$ mm), a 3D-printed middle layer (Rigid 4K, printed using a FormLab 3 \cite{Formlabs_Rigid4000}, accuracy $\approx 0.01$ mm), and an outer rubber layer (purchased from Amazon \cite{Amazon_Product}). Each layer serves a specific function. The inner layer, made of rigid material, prevents deformation from finger pressure when the sensor is worn on the hand. The middle layer, slightly more compliant, enhances strain transmission to the FBGs, improving signal-to-noise ratio. The outer rubber layer ensures effective transmission of external contact forces for continuous signal detection.

All three layers share a cylindrical body with a dome-shaped top. The optical fiber is coiled and adhered to the cylindrical body of the middle layer. The inner radius is chosen based on the operator’s finger size but remains above 8\,mm to meet the minimum bend radius of the fiber. The sensor employs Ø0.25 mm Corning Ultra SMF-28 glass fiber with an acrylate coating (Ø0.290\,mm), with Bragg Gratings evenly distributed along the sensor surface. The FBGs have nominal wavelengths from 1525 to 1565\,nm and are sampled at 2\,kHz using a Micron Optics sm130 interrogator, following the methodology in \cite{hill1997fiber}.

\begin{figure}[t]
\centering
    \includegraphics[width=0.95\linewidth]{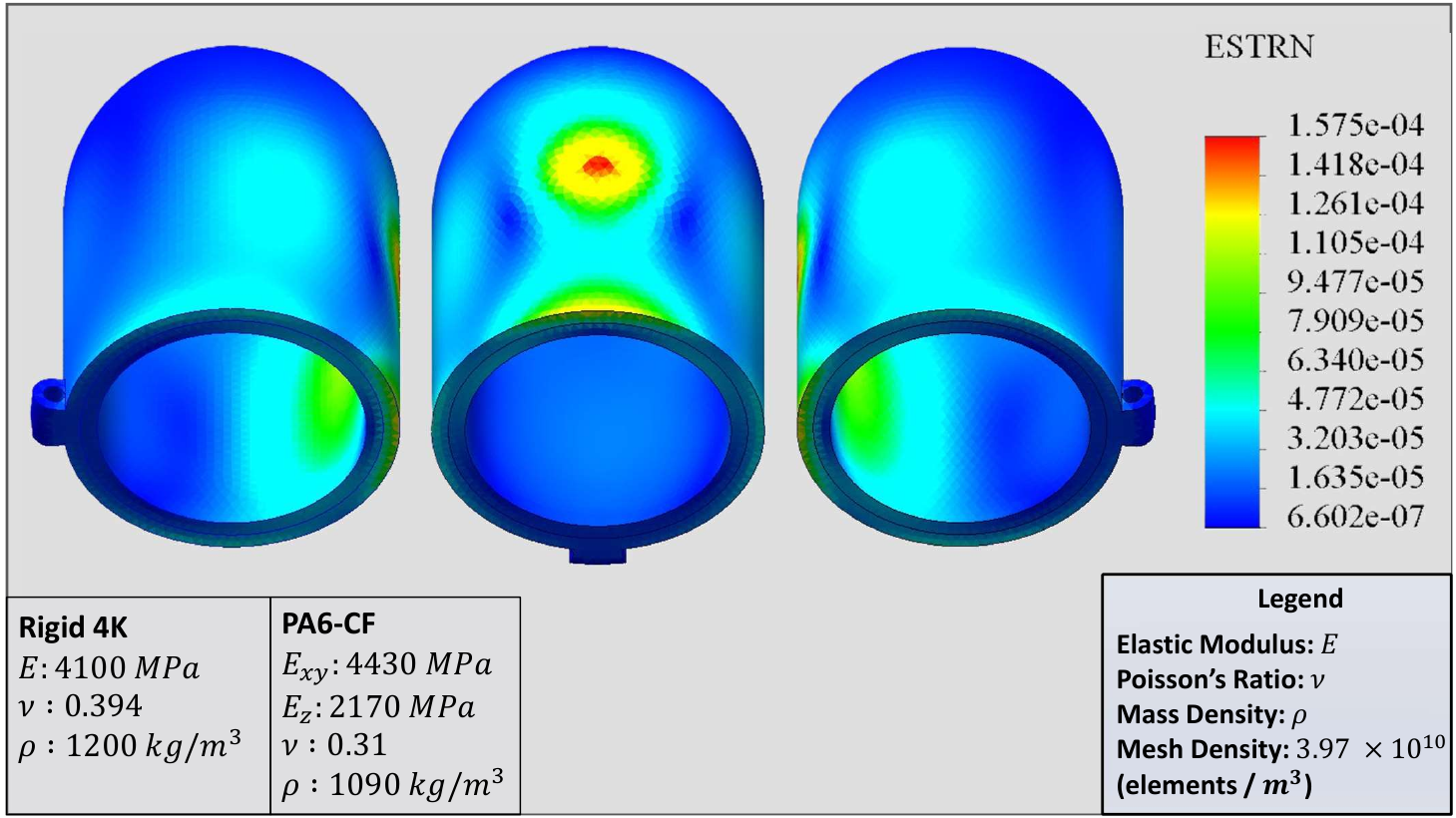}
    \caption{Finite element analysis of the tactile sensor illustrating the strain distribution across the dome structure. The color gradient represents strain magnitude, with red indicating regions of higher strain.}
\label{fea}
\vspace{-2em}
\end{figure}

Finite Element Analysis (FEA) was used to guide the placement of the FBGs. As shown in Fig.~\ref{fea}, strain changes are most pronounced near the point of applied force and decrease with distance, confirming that sensor responses are localized. Additionally, the strain distribution exhibits symmetry, reflecting the sensor’s symmetric design. Based on these results, we evenly distributed the FBGs across the sensor surface to ensure comprehensive tactile sensing coverage. While this approach provides an effective initial placement strategy, a formal optimization of the mapping from contact forces to FBG signals remains a topic for future work.

\subsection{Fabrication}
The fabrication process is straightforward. After 3D printing the inner and middle layers, the inner layer is inserted into the middle layer with a tight fit. Before assembly, the optical fiber is threaded through a Polytetrafluoroethylene (PTFE) tube to protect the excess length leading to the interrogator. The fiber is then affixed to the middle layer using Loctite Instant Adhesive 401. It is coiled around the cylindrical body, ensuring that the FBGs are positioned at designated locations. Precise placement is not critical, as fabrication uncertainties are later calibrated using our contact prediction setup. Once the fiber is secured, the outer rubber cap is fitted over the middle layer to complete the assembly. 

\vspace{-0.5em}
\section{Contact Prediction} 
Since the raw signal consists of FBG wavelength data representing strain at specific locations on the sensor structure, our objective is to extract meaningful information—specifically, the contact position. To achieve this, we assume that contact between the object and the sensor occurs at a single point. This section outlines our approach to contact localization, including the design of the calibration hardware and the development of a learning-based contact prediction algorithm. 

\subsection{Calibration System}
\begin{figure}[t]
      \centering
      \includegraphics[width=0.8\linewidth]{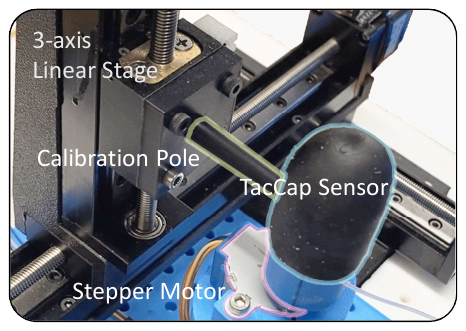}
      \caption{Calibration setup for collecting contact prediction data.}
      \label{cali}
\vspace{-2em}
\end{figure}

The goal of the calibration system is to obtain accurate contact-sensor signal pairs in a cost-effective and easily reproducible manner. As shown in Fig.~\ref{cali}, we designed a calibration probe with a hemispherical head to ensure precise point contact. The probe is mounted on a three-axis linear stage, each axis spanning 150\,mm, and driven by an L298N driver, controlled by a Teensy board. Additionally, the TacCap sensor is mounted on a rotational stage equipped with a stepper motor.
\subsection{Calibration Process}
The calibration process involves collecting paired contact data for each TacCap sensor, covering the surface regions where contact is most likely to occur. We discretize the front surface of the sensor into an $n \times m$ grid. During each calibration pass, the calibration stage makes contact with the TacCap sensor at a predefined grid point. At each contact location, multiple contact attempts with random slight displacements are applied to the soft outer layer of the sensor to account for variations. The rotational stage then rotates by an angle $\theta$. This process continues until the sensor is rotated by a total of 360 degrees, ensuring comprehensive coverage of the contact surface.

\subsection{Data Preprocessing}

We model the contact prediction problem as a regression task, so we first map the location on the TacCap to a continuous space. Given a TacCap sensor with radius \(r\) and height \(h\), the contact point \(l (i, j)\) at \((i, j)\) on the \(n \times m\) grid after \(k\) rotations is defined as:
\[
l(i,j) = 
\begin{bmatrix}
\arcsin\left(\frac{n-2i}{2r}d_x\right) + k \cdot \Delta \theta \\
j \cdot d_y
\end{bmatrix}
\]
where \(d_x\) and \(d_y\) are the grid spacing along the x and y axes, respectively, and \(\Delta \theta\) is the rotation angle increment.

\begin{figure}[b]
\centering
    \includegraphics[width=0.95\linewidth]{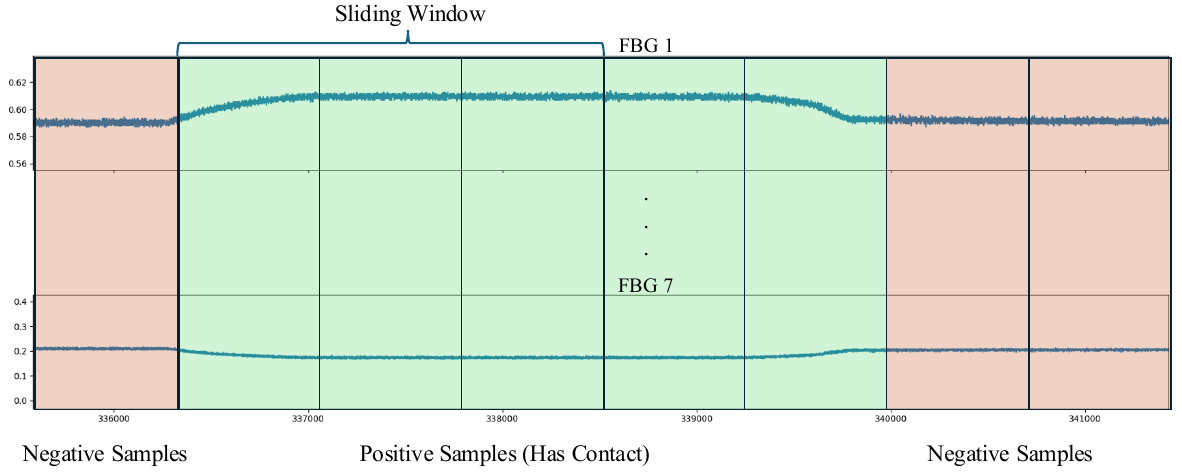}
    \caption{Illustration of the raw wavelength signal processing during calibration. A sliding window is applied to capture signal history. Signals at the right end of the sliding window, falling within the green region, are considered positive (indicating contact), while those in the red region are considered negative.}
\label{contact_raw}
\end{figure}

The FBG wavelength data is continuously recorded at 2\,kHz, as shown in Fig.~\ref{contact_raw}. Given the index corresponding to the midpoint of each contact event, denoted as \( \text{idx}_\text{mid} \), and a positive window size \( w_p \), we classify signals within the range:
\[
[\text{idx}_\text{mid} - w_p, \text{idx}_\text{mid} + w_p]
\]
as positive (indicating contact), while all other signals are labeled as negative.
To create each pair of data, we use a sliding window of length \(w_d\). The goal is to predict \(l_{\text{idx}}(i, j)\) based on FBG signals within the range \([\text{idx} - w_d, \text{idx})\). The training set and test set are chosen from different samples with no overlap.

\begin{figure}[t]
      \centering
      \includegraphics[width=0.95\linewidth]{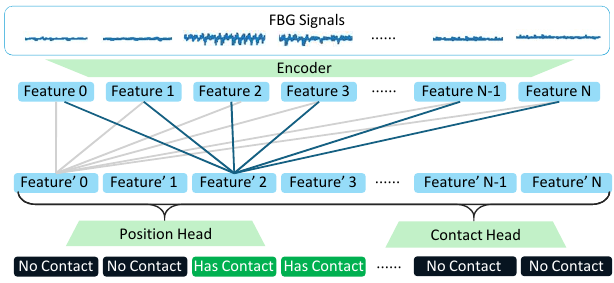}
      \caption{Architecture of the contact prediction network.}
      \label{architecture}
\vspace{-2em}
\end{figure}

\subsection{Contact Prediction Model}

To predict the contact position and determine whether contact has occurred, we design a learning-based model that processes the FBG signals through an encoder-decoder structure with multi-task learning, as shown in Fig.~\ref{architecture}.

\subsubsection{Model Architecture}

The raw FBG signals, denoted as \( S \in \mathbb{R}^{T \times \text{dim}_s} \), where \( T \) is the sequence length and \( \text{dim}_s \) is the number of sensors, are first passed through an encoder to extract contact feature representations. The encoder outputs a set of feature vectors:

\[
F = \text{Encoder}(S), \quad F \in \mathbb{R}^{T \times \text{dim}_f}
\]
where \(\text{dim}_f\) is the feature dimension. These features are then processed by a self-attention structure that captures dependencies across time steps:

\[
F' = \text{Decoder}(F), \quad F' \in \mathbb{R}^{T \times \text{dim}_f}
\]
where \( F' \) represents the refined feature representations after contextual learning. 

\subsubsection{Multi-Task Learning for Contact Prediction}

To simultaneously predict the contact position and determine the presence of contact, we employ a multi-task learning approach with two separate heads:

\textit{Position Head}: Predicts the contact position as a regression problem. Given the refined features \( F' \), the position probability distribution is:

\[
P_{\text{pos}} = \text{Head}_{\text{pos}}( F')
\]
where \( \text{Head}_{\text{pos}} \) is a linear layer.

\textit{Contact Head}: Determines whether contact has occurred using a binary classification approach:

\[
P_{\text{contact}} = \text{Head}_{\text{contact}} (F')
\]
where \(\text{Head}_{\text{contact}}\) is a linear layer.

\textit{Loss Function}: To train the model, we minimize the combined loss function:

\[
\mathcal{L} = \mathcal{L}_{\text{pos}} + \alpha \cdot \mathcal{L}_{\text{contact}},
\]

\noindent where \(\alpha\) is a weighting coefficient to balance the tasks. The positional loss \( \mathcal{L}_{\text{pos}} \), defined as the shortest possible distance between the predicted and ground truth contact positions, is given by:

\[
\mathcal{L}_{\text{pos}} = \min\left(\frac{1}{N}\sum_{i=1}^N (p_i - t_i)^2, \frac{1}{N}\sum_{i=1}^N (p_i + 1 - t_i)^2\right),
\]

\noindent where \(p_i\) and \(t_i\) are the normalized predicted and target position vectors, respectively, for samples where contact exists. \(N\) is the number of samples with contact. The contact classification loss \( \mathcal{L}_{\text{contact}} \) is computed using cross-entropy loss.

The accuracy of contact prediction is assessed using the Euclidean distance between the predicted and ground truth contact positions. A detailed discussion of the experiments and results is provided in the following section.

\begin{figure}[t]
      \centering
      \includegraphics[width=0.95\linewidth]{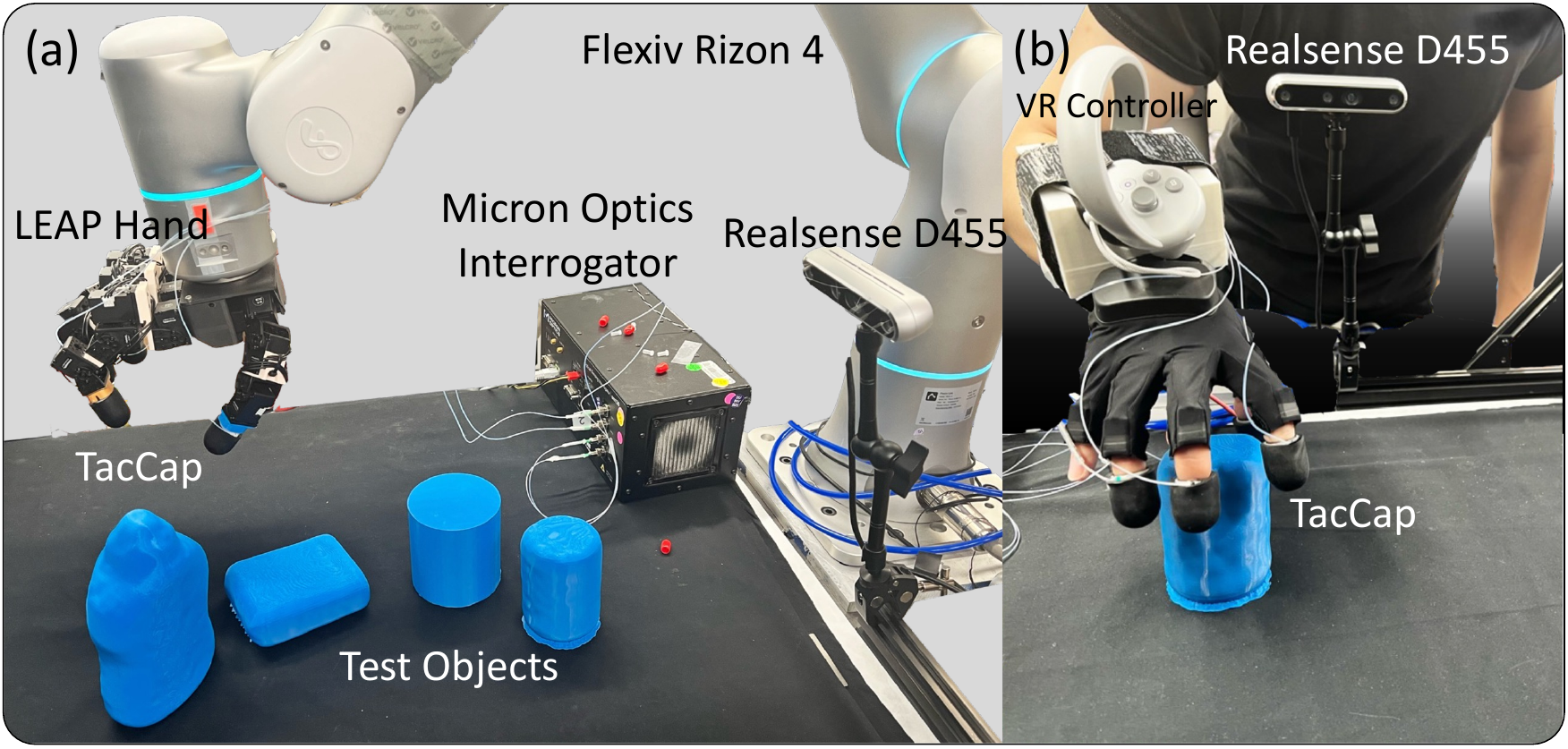}
      \caption{(a) Grasp Stability Prediction: TacCap sensors are mounted on a hand attached to a robotic arm. An optical fiber connects the sensors to an interrogator placed on the table. A chest-view camera captures the task space. (b) Grasp Stability Human Demonstration: A human operator wears the same TacCap sensors, with a VR controller attached to the wrist for tracking wrist pose. The same chest-view camera captures the task space.}
      \label{grasp_setup}
\vspace{-2em}
\end{figure}

\section{Experiments and Results}
In this section, we conduct a series of experiments to evaluate the performance of the TacCap sensor, with a particular focus on its effectiveness in human-to-robot demonstration transfer. These experiments aim to answer the following key questions:

\begin{itemize}
\item How does the sensor perform in terms of sensitivity and response time?
\item How consistent are the sensor readings over repeated use, and what are the variations across different sensors?
\item How significant is the transfer gap between human demonstration and robotic execution?
\end{itemize}

\subsection{Sensor Performance: Sensitivity and Response Time}

We evaluate the sensor using a 6-axis force-torque ATI sensor to apply controlled forces in the normal direction to the sensor surface. To assess sensitivity, we apply forces at varying magnitudes and determine the minimum detectable force threshold, defined as the smallest force at which a noticeable shift in wavelength occurs. This threshold is measured at 0.028\,N, where the signal change surpasses a predefined noise level threshold. We define this threshold as $\Delta\lambda_{\text{threshold}} = 3\sigma_{\text{baseline}}$, where $\sigma_{\text{baseline}}$ is the standard deviation of the wavelength signal under no applied force. 

Following a similar approach, we measure the rise time and fall time of the sensor response. The rise time is 87 ms, defined as the time taken for the signal to reach 90\% of the way from the initial value to the steady-state value force application. The fall time is 92 ms, representing the time required for the signal to return to 10\% of this range after force removal.

\subsection{Sensor Repeatability and Contact Prediction Consistency}
To evaluate sensor repeatability and potential degradation over time, we first use our calibration setup to contact the sensor at a fixed location and record its signal before prolonged usage. To simulate prolonged usage, we modify our setup to repeatedly contact the front surface of the sensor using the calibration stage, ensuring coverage of a large portion of potential contact points encountered during grasping. The sensor undergoes a total of 1760 contact cycles during this phase.

\begin{figure}[t]
      \centering
      \includegraphics[width=0.95\linewidth]{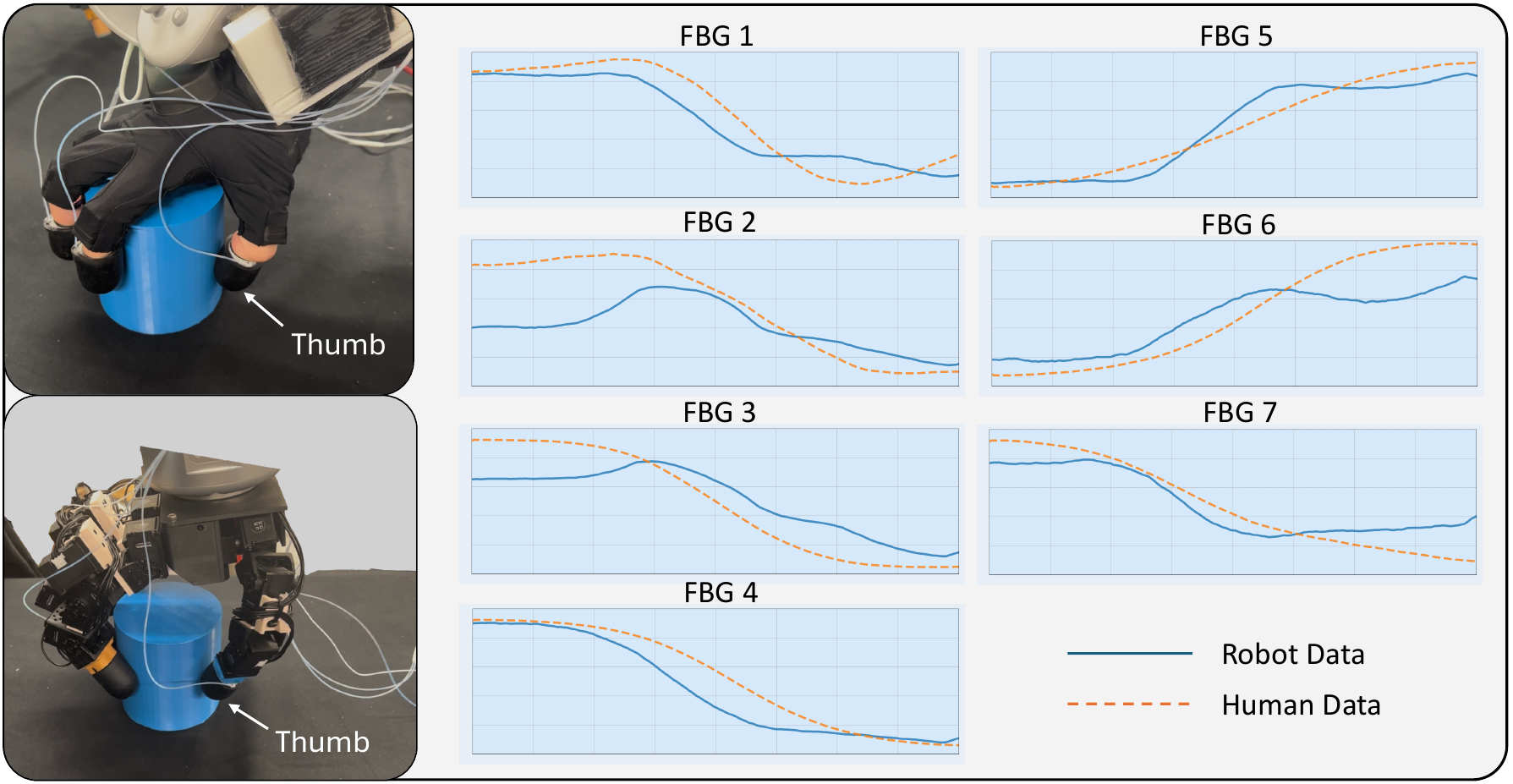}
      \caption{Raw signal data from FBG sensors on the thumb, worn on a human hand and mounted on a robotic hand, while grasping the same object. The horizontal axis represents time, and the vertical axis shows normalized wavelength signals.}
      \label{raw_data_with_demo}
\vspace{-1em}
\end{figure}

After completing the prolonged usage phase, we again touch the original fixed location on the sensor and record its FBG signal. The absolute wavelength difference before and after long-term use is calculated as:

\[
    \Delta{\text{degradation}} = 1- \frac{R_{\text{initial}}}{R_{\text{after use}}} %
\]
where \(R_{\text{initial}}\) represents the sensor’s wavelength range in multiple contacts before prolonged use, and \(R_{\text{after use}}\) is the range recorded after repeated touches. The calculated average degradation difference is {\(3.74\%\). The difference of the DIGIT sensor is \(14\%\) after 100 repeated touches. 

To assess contact prediction consistency across different sensor instances, we compare the contact localization accuracy among three independently fabricated TacCap sensors. Identical contact prediction tests are performed on all sensors using the same set of predefined touch locations. The accuracy of contact prediction is evaluated based on the Euclidean distance between the predicted contact position and the ground truth. The average prediction error for each sensor is summarized in Table~\ref{tab:contact_accuracy}.

\begin{table}[t]
    \centering
    \caption{Contact Prediction Consistency}
    \begin{tabular}{c|ccc}
        \hline
        & Sensor 1 & Sensor 2 & Sensor 3\\
        \hline
        Accuracy (mm) & 5.3 & 5.8 & 5.36 \\
        \hline
    \end{tabular}
    \label{tab:contact_accuracy}
    \vspace{-2em}
\end{table}

\subsection{Grasp Stability Prediction}
To evaluate the human-to-robot transfer gap, we conduct grasp stability prediction experiments using a procedure similar to that described in \cite{si2022grasp} and a similar hardware setup to \cite{wang2024dexcap}. The task is formulated as a supervised learning problem, where we use tactile signals recorded during grasping to predict actions. Grasp stability is classified into binary labels: a successful grasp is defined as one where the object can be stably lifted, while a failure occurs if the object slips or is dropped.

\subsubsection{Hardware Setup}
Our experimental setup is illustrated in Fig.~\ref{grasp_setup}. We selected four objects with varying geometric shapes. Instead of using a wrist-mounted camera as in \cite{wang2024dexcap}, we utilize a Pico VR headset to track wrist poses and a chest-view camera to capture the task space. To minimize the visual domain gap, we use the same camera for data collection when wearing TacCap. For hand pose capture, we use a Rokoko motion capture glove to record 20 successful grasp demonstrations. For comparison, we use the same setup to teleoperate the robot and collect demonstration data with our TacCap sensor and GelSight DIGIT sensor.

As an initial validation step, we visualize raw signals collected from both wearable and teleoperation-based demonstrations (Fig.~\ref{raw_data_with_demo}). This visualization highlights the alignment of signals between the human hand and the robotic hand.

\subsubsection{Task Setup}
The task setup involves placing an object randomly within a small range on a table while the robotic arm starts from a fixed position in the air. We designed two tasks to evaluate the performance in both wearable transfer and teleoperation, which are grasp stability tasks and a skill learning task.

In grasp stability tasks, we initialize the robotic hand directly above the object using a heuristic policy. In this setup, the model is trained solely on tactile and proprioception data, predicting only the grasping action required to lift and hold the object stably. 

In the skill learning task, the robot must adjust its pose, align with the object, descend to grasp it, and lift it. The model is trained with fused feature of both vision and tactile information. 

We train separate models based on these demonstrations and evaluate it by performing 10 grasp attempts per object to assess the transfer gap. The success rate (SR) is defined as the number of successful grasps divided by the total number of attempts.

\subsubsection{Model Architecture and Training}
We employ a standard ResNet-18 model to encode visual data, while tactile signals and proprioception data are processed through a set of linear layers. A linear prediction head is then used to predict the grasping action. The action space is defined by the joint angles of the LEAP Hand \cite{shaw2023leap} and the 6-axis position of the robot arm's end. All data are collected at 30 FPS. We downsample the data by a factor of 3 and use a window length of 30 to predict the next action.

\subsection{Grasp Stability Results}
Table \ref{results} summarizes our experimental results. We conduct ablation studies across these scenarios:
\begin{itemize}
    \item Training with demonstration data collected using our wearable setup, with and without TacCap sensor data.
    \item Training with demonstration data collected using teleoperation, with and without TacCap sensor data.
    \item Training with a mixed dataset containing 10 teleoperation and 10 wearable setup demonstrations.
    \item Training with teleoperation-collected data using the GelSight DIGIT sensor.
\end{itemize}

\subsubsection{Grasp Stability Results}
As shown in the left parts of Table \ref{results}, the performance of data collected via wearable and teleoperation methods is similar, showing the low transfer gap. This is largely attributed to our design, where the sensors are mounted in a similar manner on both the human and robotic hands, resulting in nearly identical contact locations during grasping. The low transfer gap is also qualitatively demonstrated in Fig.~\ref{raw_data_with_demo}, which shows the raw FBG signal from both data collection methods.

Comparing scenarios with and without the TacCap sensor demonstrates that incorporating TacCap sensor data significantly improves performance in both wearable and teleoperation settings. Interestingly, the wearable setup experiences a greater drop in success rate than teleoperation when the TacCap sensor is not used. This is likely due to errors in the Rokoko glove hand capture, which create a small gap between the robot’s fingers and the object. However, when the TacCap sensor is used, the model encourages the hand to grasp the object more tightly, mitigating this issue.

\subsubsection{Skill Learning Results}
To further demonstrate that our TacCap sensor improves grasp stability in a policy fused with vision, the results shown in the right part of Table \ref{results} indicate that models trained with tactile sensor data perform better than those trained without it. A performance drop is observed when training with data collected from the wearable setup, primarily due to differences in visual data. However, incorporating a few teleoperation data significantly improves performance by reducing the domain gap.

Additionally, training with teleoperation data using the DIGIT sensor yields lower performance compared to using our TacCap sensor. This is likely because the TacCap sensor can sense 360 degrees around the fingertip, whereas the DIGIT sensor can only sense the front. In practical grasping scenarios, we observed that touch on the sides is very common. Furthermore, we noticed that the high-dimensional nature of the DIGIT sensor data may confuse the network, whereas the more straightforward signals from the TacCap sensor, which directly correspond to grasp force, are easier for the model to learn.
\vspace{-1em}

\begin{table}[h]
    \centering
    \caption{Grasp Stability Prediction (Left) and without Vision (Right) \\ success rates (SR)}
    \begin{minipage}{0.48\linewidth}
        \centering
        \begin{tabular}{c|c}
            \hline
            Condition & SR (\%) \\
            \hline
            Wear w/ TacCap & 90 \\
            Wear w/o TacCap & 10 \\
            Teleop w/ TacCap & 100 \\
            Teleop w/o TacCap & 80 \\
            \hline
        \end{tabular}
    \end{minipage}
    \hfill    
    \begin{minipage}{0.48\linewidth}
        \centering
        \begin{tabular}{c|c}
            \hline
            Condition & SR (\%) \\
            \hline
            Wear w/ TacCap & 10 \\
            Wear w/o TacCap & 0 \\
            Teleop w/ TacCap & 75 \\
            Teleop w/o TacCap & 60 \\
            Mixed w/ TacCap & 50 \\
            Teleop w/ DIGIT & 40 \\
            \hline
        \end{tabular}
    \end{minipage}
    \label{results}
\vspace{-1.5em}
\end{table}
\section{Conclusion and future work}
In this paper, we introduced TacCap, a wearable FBG-based tactile sensor designed for seamless human-to-robot skill transfer. TacCap is lightweight, durable, and immune to environmental interference, making it well-suited for real-world tactile sensing applications. We detailed its design, fabrication, and calibration, demonstrating its high sensitivity, fast response time, and repeatability. To evaluate its effectiveness, we conducted contact prediction experiments, showing that TacCap can localize contact points on the sensor surface. Additionally, we integrated TacCap into a grasp stability prediction task, performed an ablation study, and compared its performance with the GelSight DIGIT sensor. The results demonstrate that incorporating tactile sensing improves grasp success rates and the human-to-robot transfer gap is low.

While our results show the potential of TacCap for human-to-robot skill transfer, several limitations remain. First, the accuracy of our contact prediction model can be further improved by refining our sensor calibration process and releasing the assumption of single contact to deal with more complicated cases. We also aim to explore contact prediction in more sophisticated manipulation tasks. Furthermore, while TacCap has been validated in structured lab environments, its robustness and scalability in dynamic, real-world conditions remain an open question. In future work, we plan to deploy TacCap for in-the-wild human demonstrations, testing its durability, usability, and effectiveness across diverse manipulation tasks and environments. Expanding our tasks with naturalistic tactile interactions will also allow us to explore more advanced multimodal learning approaches, improving the generalization of human-to-robot skill transfer.

By addressing these challenges, we aim to further bridge the gap between human and robotic tactile sensing, enabling more dexterous and adaptive robotic manipulation in real-world applications.
\vspace{-2px}
\section{Acknowledgement}
We are grateful for support from the Zhulong Innovation Fellowship and additional support from TRI Global. We would like to thank Albert Wu, Sirui Chen, Chen Wang for providing help and feedback. We especially thank Kenneth Shaw, Ananye Agrawal, Deepak Pathak for open-sourcing the LEAP Hand.
\vspace{-1em}
\bibliographystyle{ieeetran_mod}
\bibliography{ref}

\begin{thebibliography}{10}
\providecommand{\url}[1]{#1}
\csname url@rmstyle\endcsname
\providecommand{\newblock}{\relax}
\providecommand{\bibinfo}[2]{#2}
\providecommand\BIBentrySTDinterwordspacing{\spaceskip=0pt\relax}
\providecommand\BIBentryALTinterwordstretchfactor{4}
\providecommand\BIBentryALTinterwordspacing{\spaceskip=\fontdimen2\font plus
\BIBentryALTinterwordstretchfactor\fontdimen3\font minus \fontdimen4\font\relax}
\providecommand\BIBforeignlanguage[2]{{%
\expandafter\ifx\csname l@#1\endcsname\relax
\typeout{** WARNING: IEEEtran.bst: No hyphenation pattern has been}%
\typeout{** loaded for the language `#1'. Using the pattern for}%
\typeout{** the default language instead.}%
\else
\language=\csname l@#1\endcsname
\fi
#2}}

\bibitem{lee2020making}
M.~A. Lee, Y.~Zhu, P.~Zachares, \emph{et~al.}, ``Making sense of vision and touch: Learning multimodal representations for contact-rich tasks,'' \emph{IEEE Transactions on Robotics}, vol.~36, no.~3, pp. 582--596, 2020.

\bibitem{li2023see}
H.~Li, Y.~Zhang, J.~Zhu, \emph{et~al.}, ``See, hear, and feel: Smart sensory fusion for robotic manipulation,'' in \emph{Conference on Robot Learning}.\hskip 1em plus 0.5em minus 0.4em\relax PMLR, 2023, pp. 1368--1378.

\bibitem{gao2023sonicverse}
R.~Gao, H.~Li, G.~Dharan, \emph{et~al.}, ``Sonicverse: A multisensory simulation platform for embodied household agents that see and hear,'' in \emph{2023 IEEE International Conference on Robotics and Automation (ICRA)}.\hskip 1em plus 0.5em minus 0.4em\relax IEEE, 2023, pp. 704--711.

\bibitem{gao2023objectfolder}
R.~Gao, Y.~Dou, H.~Li, \emph{et~al.}, ``The objectfolder benchmark: Multisensory learning with neural and real objects,'' in \emph{Proceedings of the IEEE/CVF Conference on Computer Vision and Pattern Recognition}, 2023, pp. 17\,276--17\,286.

\bibitem{liu2024maniwav}
Z.~Liu, C.~Chi, E.~Cousineau, \emph{et~al.}, ``Maniwav: Learning robot manipulation from in-the-wild audio-visual data,'' in \emph{8th Annual Conference on Robot Learning}, 2024.

\bibitem{duan2017one}
Y.~Duan, M.~Andrychowicz, B.~Stadie, \emph{et~al.}, ``One-shot imitation learning,'' \emph{Advances in neural information processing systems}, vol.~30, 2017.

\bibitem{brohan2023rt}
A.~Brohan, N.~Brown, J.~Carbajal, \emph{et~al.}, ``Rt-2: Vision-language-action models transfer web knowledge to robotic control,'' \emph{arXiv preprint arXiv:2307.15818}, 2023.

\bibitem{chi2023diffusion}
C.~Chi, Z.~Xu, S.~Feng, \emph{et~al.}, ``Diffusion policy: Visuomotor policy learning via action diffusion,'' \emph{The International Journal of Robotics Research}, p. 02783649241273668, 2023.

\bibitem{chi2024universal}
C.~Chi, Z.~Xu, C.~Pan, \emph{et~al.}, ``Universal manipulation interface: In-the-wild robot teaching without in-the-wild robots,'' \emph{arXiv preprint arXiv:2402.10329}, 2024.

\bibitem{wang2024dexcap}
C.~Wang, H.~Shi, W.~Wang, \emph{et~al.}, ``Dexcap: Scalable and portable mocap data collection system for dexterous manipulation,'' \emph{arXiv preprint arXiv:2403.07788}, 2024.

\bibitem{sundaram2019learning}
S.~Sundaram, P.~Kellnhofer, Y.~Li, \emph{et~al.}, ``Learning the signatures of the human grasp using a scalable tactile glove,'' \emph{Nature}, vol. 569, no. 7758, pp. 698--702, 2019.

\bibitem{delpreto2022actionsense}
J.~DelPreto, C.~Liu, Y.~Luo, \emph{et~al.}, ``Actionsense: A multimodal dataset and recording framework for human activities using wearable sensors in a kitchen environment,'' \emph{Advances in Neural Information Processing Systems}, vol.~35, pp. 13\,800--13\,813, 2022.

\bibitem{luo2024adaptive}
Y.~Luo, C.~Liu, Y.~J. Lee, \emph{et~al.}, ``Adaptive tactile interaction transfer via digitally embroidered smart gloves,'' \emph{Nature communications}, vol.~15, no.~1, p. 868, 2024.

\bibitem{zhang2018deep}
T.~Zhang, Z.~McCarthy, O.~Jow, \emph{et~al.}, ``Deep imitation learning for complex manipulation tasks from virtual reality teleoperation,'' in \emph{2018 IEEE international conference on robotics and automation (ICRA)}.\hskip 1em plus 0.5em minus 0.4em\relax Ieee, 2018, pp. 5628--5635.

\bibitem{wang2023mimicplay}
C.~Wang, L.~Fan, J.~Sun, \emph{et~al.}, ``Mimicplay: Long-horizon imitation learning by watching human play,'' \emph{arXiv preprint arXiv:2302.12422}, 2023.

\bibitem{shaw2023videodex}
K.~Shaw, S.~Bahl, and D.~Pathak, ``Videodex: Learning dexterity from internet videos,'' in \emph{Conference on Robot Learning}.\hskip 1em plus 0.5em minus 0.4em\relax PMLR, 2023, pp. 654--665.

\bibitem{zhao2023learning}
T.~Z. Zhao, V.~Kumar, S.~Levine, \emph{et~al.}, ``Learning fine-grained bimanual manipulation with low-cost hardware,'' \emph{arXiv preprint arXiv:2304.13705}, 2023.

\bibitem{fu2024mobile}
Z.~Fu, T.~Z. Zhao, and C.~Finn, ``Mobile aloha: Learning bimanual mobile manipulation with low-cost whole-body teleoperation,'' \emph{arXiv preprint arXiv:2401.02117}, 2024.

\bibitem{wu2024gello}
P.~Wu, Y.~Shentu, Z.~Yi, \emph{et~al.}, ``Gello: A general, low-cost, and intuitive teleoperation framework for robot manipulators,'' in \emph{2024 IEEE/RSJ International Conference on Intelligent Robots and Systems (IROS)}.\hskip 1em plus 0.5em minus 0.4em\relax IEEE, 2024, pp. 12\,156--12\,163.

\bibitem{chen2024arcap}
S.~Chen, C.~Wang, K.~Nguyen, \emph{et~al.}, ``Arcap: Collecting high-quality human demonstrations for robot learning with augmented reality feedback,'' \emph{arXiv preprint arXiv:2410.08464}, 2024.

\bibitem{park2024dexhub}
Y.~Park, J.~S. Bhatia, L.~Ankile, \emph{et~al.}, ``Dexhub and dart: Towards internet scale robot data collection,'' \emph{arXiv preprint arXiv:2411.02214}, 2024.

\bibitem{battaglia2014thimblesense}
E.~Battaglia, G.~Grioli, M.~G. Catalano, \emph{et~al.}, ``Thimblesense: an individual-digit wearable tactile sensor for experimental grasp studies,'' in \emph{2014 IEEE International Conference on Robotics and Automation (ICRA)}.\hskip 1em plus 0.5em minus 0.4em\relax IEEE, 2014, pp. 2728--2735.

\bibitem{battaglia2015thimblesense}
E.~Battaglia, M.~Bianchi, A.~Altobelli, \emph{et~al.}, ``Thimblesense: a fingertip-wearable tactile sensor for grasp analysis,'' \emph{IEEE transactions on haptics}, vol.~9, no.~1, pp. 121--133, 2015.

\bibitem{altobelli2016wearable}
A.~Altobelli and A.~Altobelli, ``Wearable approach: Thimblesense, a fingertip-wearable tactile sensor for grasp analysis,'' \emph{Haptic Devices for Studies on Human Grasp and Rehabilitation}, pp. 43--55, 2016.

\bibitem{qiao2017fiber}
X.~Qiao, Z.~Shao, W.~Bao, \emph{et~al.}, ``Fiber bragg grating sensors for the oil industry,'' \emph{Sensors}, vol.~17, no.~3, p. 429, 2017.

\bibitem{park2011real}
S.~Park, T.~Park, and K.~Han, ``Real-time monitoring of composite wind turbine blades using fiber bragg grating sensors,'' \emph{Advanced Composite Materials}, vol.~20, no.~1, pp. 39--51, 2011.

\bibitem{presti2020fiber}
D.~L. Presti, C.~Massaroni, C.~S.~J. Leitao, \emph{et~al.}, ``Fiber bragg gratings for medical applications and future challenges: A review,'' \emph{Ieee Access}, vol.~8, pp. 156\,863--156\,888, 2020.

\bibitem{heo2006tactile}
J.-S. Heo, J.-H. Chung, and J.-J. Lee, ``Tactile sensor arrays using fiber bragg grating sensors,'' \emph{Sensors and Actuators A: Physical}, vol. 126, no.~2, pp. 312--327, 2006.

\bibitem{frishman2021multi}
S.~Frishman, J.~Di, Z.~Karachiwalla, \emph{et~al.}, ``A multi-axis fbg-based tactile sensor for gripping in space,'' in \emph{2021 IEEE/RSJ International Conference on Intelligent Robots and Systems (IROS)}.\hskip 1em plus 0.5em minus 0.4em\relax IEEE, 2021, pp. 1794--1799.

\bibitem{massari2022functional}
L.~Massari, G.~Fransvea, J.~D’Abbraccio, \emph{et~al.}, ``Functional mimicry of ruffini receptors with fibre bragg gratings and deep neural networks enables a bio-inspired large-area tactile-sensitive skin,'' \emph{Nature Machine Intelligence}, vol.~4, no.~5, pp. 425--435, 2022.

\bibitem{li2024whisker}
H.~Li, C.~Xing, S.~Khan, \emph{et~al.}, ``Whisker-inspired tactile sensing: A sim2real approach for precise underwater contact tracking,'' \emph{arXiv preprint arXiv:2410.14005}, 2024.

\bibitem{BambuLab_PA6CF}
\BIBentryALTinterwordspacing
{Bambu Lab}, ``Pa6-cf filament,'' 2025, accessed: 2025-02-26. [Online]. Available: \url{https://us.store.bambulab.com/products/pa6-cf?srsltid=AfmBOoocwBZQ9TBJ6sKRYWqYbfMZ__45mSoWDBm2OP8AN9pn2948ooHU}
\BIBentrySTDinterwordspacing

\bibitem{Formlabs_Rigid4000}
\BIBentryALTinterwordspacing
{Formlabs}, ``Rigid 4000 resin,'' 2025, accessed: 2025-02-26. [Online]. Available: \url{https://formlabs.com/store/materials/rigid-4000-resin/}
\BIBentrySTDinterwordspacing

\bibitem{Amazon_Product}
\BIBentryALTinterwordspacing
{Amazon}, ``Amazon product page,'' 2025, accessed: 2025-02-26. [Online]. Available: \url{https://a.co/d/8F2Xdk1}
\BIBentrySTDinterwordspacing

\bibitem{hill1997fiber}
K.~O. Hill and G.~Meltz, ``Fiber bragg grating technology fundamentals and overview,'' \emph{Journal of lightwave technology}, vol.~15, no.~8, pp. 1263--1276, 1997.

\bibitem{si2022grasp}
Z.~Si, Z.~Zhu, A.~Agarwal, \emph{et~al.}, ``Grasp stability prediction with sim-to-real transfer from tactile sensing,'' in \emph{2022 IEEE/RSJ International Conference on Intelligent Robots and Systems (IROS)}.\hskip 1em plus 0.5em minus 0.4em\relax IEEE, 2022, pp. 7809--7816.

\bibitem{shaw2023leap}
K.~Shaw, A.~Agarwal, and D.~Pathak, ``Leap hand: Low-cost, efficient, and anthropomorphic hand for robot learning,'' \emph{arXiv preprint arXiv:2309.06440}, 2023.

\end{thebibliography}
\end{document}